%% file: main.tex
\pdfoutput=1
\documentclass[sigconf,authorversion]{acmart}
\makeatletter
\renewcommand\@formatdoi[1]{\ignorespaces}
\makeatother

\usepackage[ruled,vlined,linesnumbered]{algorithm2e}

\usepackage{multirow}
\usepackage{flushend}
\usepackage{amsmath,amssymb}
\usepackage{mathtools}
\usepackage[acronym]{glossaries}
\usepackage{units}
\usepackage{booktabs}
\usepackage[inline]{enumitem} % Inline lists
\usepackage{subcaption} %  for subfigures environments 
\usepackage{hyphenat}

\setcopyright{none}
\acmConference[REVEAL '19]{the ACM RecSys Workshop on Reinforcement Learning and Robust Estimators for Recommendation Systems}{September 20th, 2019}{Copenhagen, Denmark}
\acmYear{2019}
\copyrightyear{2019}

\DeclareMathOperator*{\E}{\mathbb{E}}

\begin{document}
\title[]{Ranking metrics on non-shuffled traffic}

\author{Alexandre Gilotte}
\affiliation{
  \institution{Criteo AI Lab}
  \city{Paris}
  \country{France}
}
\email{a.gilotte@criteo.com}

\renewcommand{\shortauthors}{A. Gilotte et al.}

\begin{abstract}
Ranking metrics are a family of metrics largely used to evaluate recommender systems. However they typically suffer from the fact the reward is affected by the order in which recommended items are displayed to the user. A classical way to overcome this position bias is to uniformly shuffle a proportion of the recommendations, but this method may result in a bad user experience. It is nevertheless common to use a stochastic policy to generate the recommendations, and we suggest a new method to overcome the position bias, by leveraging the stochasticity of the policy used to collect the dataset.
\end{abstract}


\maketitle

\input{body}

\bibliographystyle{ACM-Reference-Format}
\bibliography{bibliography}

\end{document}

%% file: body.tex
\section{Introduction}

One of the main challenges in online advertising is to select products relevant to the user in each displayed banner. 
The choice of those products is typically made by sophisticated recommender systems, which are optimized to maximize business metrics such as number of clicks or number of sales.
This is typically done by scoring each of the available products, for example predicting an expected reward per product, and then displaying the top scored products. A randomized policy is often used to select and order those top products to keep some diversity, both for the user and for training the next models. \\
The list of selected product is then shown to the user, who may click on one of the products of the banner.
\\ 

Building and improving the scoring function requires a lot of iterations between different versions, 
and therefore requires to be able to compare their performances.
This can be done by A/B testing: deploying the new version on a subset of the users, and comparing the business performances to a reference set of users.
However, gathering enough data to get statistically significant results on the performances of a test version requires to allocate an important part of the traffic, and may prove costly when the test version performs poorly.
As a consequence, it is not a practical solution to compare a large number of possible versions of the system, and instead offline metrics are used to evaluate from past logged data the performances of a new model. These offline metrics may be sorted in several families:

\begin{itemize}
\sloppy% to prevent underfull hboxes -- https://latex.org/forum/viewtopic.php?t=8752
\nohyphens{
\item  Point-wise metrics estimate the error made by a model predicting expected reward on each product. Typical example in this family is the mean square error. However, those metrics do not take into account that the model is actually used to rank the products.
\item Counterfactual metrics, which use knowledge of the stochastic policy to estimate the expected value of a business metric (such as the number of clicks) if we were using the test model, with an importance weighting scheme. More details on those methods may be found in \cite{BottouCounterfactual} , \cite{Swaminathan/Joachims/15b} \cite{gilotteoabt}. While a promising field of research, they typically suffer from a high variance which limits their use. 
\item Ranking metrics compare the ordering of the displayed products according to a scoring function $score(p)$ with the partial ordering from the user feedback (For example, when using 'click' as feedback, this partial ordering is defined by "a clicked product > a non-clicked product"). A common shortcoming of those metrics is that the user feedback can be strongly influenced by the ordering which was used to show the recommendations to the user. Typically, items proposed first by the production system have their likelihood to receive a click increased. This phenomenon, known as the position bias, has already received a lot of attention and several authors proposed solutions to try to minimize it. Work on those topics include \cite {Maarten_IRMetrics}, \cite{Chapelle_DynamicBayesianNetwork2009} and \cite{Rendle_DBLP}, or more recently \cite{Liang2016ModelingUser}, \cite{Wang2018PositionBE}, \cite{Schnabel2016RecommendationsAT},   \cite{Agarwal2018CounterfactualLF} or \cite{Joachims2017UnbiasedLW}.
}
\end{itemize}

The metric we propose in this work belongs to the family of ranking metrics, but use the knowledge of the stochastic policy to remove the position bias in a novel way.
In section 2, we focus on the pairwise agreement metric and  re-examine why it is affected by the position bias.
In section3, we propose a modified version of pairwise disagreement, which avoids position bias by using the logging policy to sample the negative product. 
In section 4, we show experimental results on Criteo data, suggesting that the proposed metric could advantageously replace the usual ranking metric, at least in this setup.
Finally, in section 5, we discuss some possible extensions along the same idea.
 
\section{Pairwise disagreement and position bias }
 
\subsection{Setting and Notations}

We consider a recommender system in the context of online advertising. The recommender system receive some query $x$, describing the context of a banner (like the size of the banner or the user history) and a list of potential products to display in the banner. Each query comes from an unknown distribution $P_x$, and the queries are supposed iid.  \\
On each query, the system should select a subset $\{p_1, p_2, ..., p_n\}$ of constant size $n$ of the set of candidates, and an ordering $\omega$ of those products. \\
The banner is then displayed to the user with product $\omega(1)$ at rank $1$ , $\omega(2)$ at rank $2$ , etc...
% $\omega \in S_n$  is a permutation of the set $\{1,...n\} $ . ( \omega(1)$ is the product at rank $1$ etc... )\\

We will assume that the production system is choosing the ordering with a stochastic policy, and define $\pi(\omega ) $ as the probability that production system choose the ordering $\omega$ on a context $x$ after choosing the set of products $\{P_1, P_2, ..., P_n\}$ (We are here omitting $x$ and $\{P_1, P_2, ..., P_n\}$ in this notation to keep it concise.)
\\
The system then receive a feedback $Y$ of the user, a binary vector where $Y_i = 1$ indicate that the user interacted (for example, clicked) with product $i$. We will also assume for simplicity that at most one of the $y_i$ is non zero, and note $R(Y)$, or simply $R$, the position in the banner of the clicked item, if any, and note this event $R>0$ . (In which case, with our notations, $Y_{\omega(R)} = 1 $ ).
\\
We also have a test model $\sigma$ scoring each product, which we want to evaluate.
\\
%We will also define $p_i \succ_\s p_j $ as 1 if  $\s(i) > \s(j)$ ,  else 0, and note $U_n$ the uniform discrete distribution on the set $\{1 ... n\}$ elements, and $U(S_n)$ the uniform distribution on the set of permutations of n elements.

\subsection{Pairwise disagreement}

One of the simplest example of ranking metric is the pairwise disagreement, which can be defined as the proportion of pairs of products which are ordered differently by the evaluated model and by the user feedback (among the pairs comparable by both orders). \\

It is usually defined, for one banner with a reward $y>0$, as: 
$$PD( s , y) := \frac{ \sum\limits_{i,j \in (1,...n)}  [\sigma(i) < \sigma(j) ][y_i = 1][y_j = 0] }{ \sum\limits_{i,j \in (1,...n)} [\sigma(i) \neq \sigma(j) ] [y_i = 1][y_j = 0] }   $$

Its value on the dataset can be computed as the expected result of the following algorithm, conditional to the fact that the sample is not rejected:
\begin{algorithm}
\DontPrintSemicolon
 \caption{Get one sample of pairwise disagreement \label{alg:sample1} }
     \SetAlgoLined
     Sample one banner from the dataset\;
     Reject it if there is not at least one clicked and one non clicked product\;
     Let $P_+$ the clicked product from this banner\;
     Sample a product $P_-$ uniformly from the non clicked products of the same banner\;
     If $score(P_+) = score(P_-) $ , reject the sample\;
     Else return $\mathbf{1}_{score(P_+) < score(P_-)} $ \;
\end{algorithm}

A perfect model would here get a pairwise disagreement of 0, while a random model would get $0.5$.

\subsection{Position bias}

As stated in the introduction, the order in which recommended items are presented to the user may have an important impact on the feedback of this user. In many systems, user is much more likely to click on the first item than on the next ones. \\
This effect induces a bias in the ranking metric: products commonly displayed in top position by the system are more likely to get clicked, just because they are in top position, not necessarily because they are really better products. This can have a very annoying effect on the pairwise disagreement metric: the ranking minimizing the expectation of the pairwise disagreement is not necessarily the model placing the 'best' product at the first position.
\\
Prior work on position bias (sometimes called "bias of rank") includes \cite {Maarten_IRMetrics,Chapelle_DynamicBayesianNetwork2009,Rendle_DBLP,Liang2016ModelingUser,Wang2018PositionBE,Schnabel2016RecommendationsAT,Agarwal2018CounterfactualLF,Joachims2017UnbiasedLW}
 %%This phenomenon, known as the position bias, has already received a lot of attention and several authors proposed solutions to try to minimize it. Work on those topics include \cite {Maarten_IRMetrics,Chapelle_DynamicBayesianNetwork2009,Rendle_DBLP}, or more recently \cite{Liang2016ModelingUser,Wang2018PositionBE,Schnabel2016RecommendationsAT,Agarwal2018CounterfactualLF,Joachims2017UnbiasedLW}.
\\

We would like here to outline why this position bias is happening:
With notations of algorithm~\ref{alg:sample1}, for two  products $p_1$ and $p_2$ such that the proportion of observed pairs $(P_+=p_1, P_-=p_2)$ is higher than the proportion of pairs in the reverse order $(P_+ = p_2, P_- = p_1)$, pairwise disagreement rewards the model for scoring $p_1$ higher than $p_2$.\\
The cause of "position bias" is that those proportions may differ not because of the relative quality of products $p_1$ and $p_2$, but because $\pi$ put one product on a position likely to get clicked more often than the other.
% This intuitive idea is made more formal in appendix~\ref{annexeExpectation}.
\\

\subsection{Existing methods to deal with position bias}

\label{ShufflingBanners}
To overcome this problem, the gold standard is to uniformly shuffle the banner before displaying it to the user. This ensure that for any rank $r$, each product has the same probability of being displayed at rank $r$, and thus if a product receives more clicks on average that another, it cannot be explained by position bias. However, uniform shuffling is often not possible or too costly to use. \\
Other methods have been proposed to limit this effect:
\begin{itemize}
\item Modeling jointly the effect of position and the effect of the recommended products. Some recent advance in this direction includes \cite{Liang2016ModelingUser, Wang2018PositionBE}.
\item Using some importance weighting scheme to simulate shuffled data was proposed by \cite{Schnabel2016RecommendationsAT}, and developed in  \cite{Agarwal2018CounterfactualLF, Joachims2017UnbiasedLW}.
\end{itemize}
However, those methods typically rely on a model and/or on some additional assumptions on how the rank affect the user feedback. A possible downside of relying of such assumptions is that the metric might under-evaluate a scoring function which would be trained with different assumptions on the effect of rank.

\section{Changing the sampling of the products in the metric  }

\subsection{Counterfactual disagreement}

To compare each pair of product $(p_1,p_2)$ more fairly, we would like to ensure that, if products $p_1$ and $p_2$ have the same click through rate when they are displayed at rank $r$, then :
$\mathbb{P}(P_+ = p_1, P_- = p_2, R = r ) = \mathbb{P}(P_+ = p_2, P_- = p_1 , R = r ) $ 

\begin{figure}[ht!]
\includegraphics[width=0.4\linewidth]{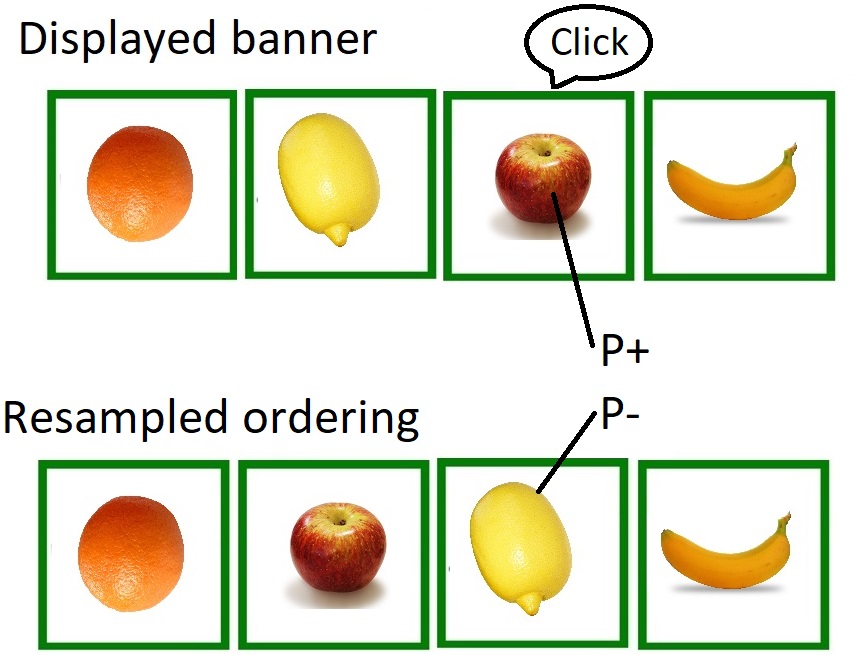}
\caption{\label{fig:banner_and_ressample} Comparing clicked product to the product at the same rank in the resampled banner }
\end{figure}

One way to ensure that is to sample a second ordering from $\pi$, and to set the negative product $P_-$ as the product whose rank in this second ordering is the same as the rank of the clicked product in the displayed ordering, as illustrated in figure~\ref{fig:banner_and_ressample}.

We thus define the "Counterfactual disagreement" as the expectation of the result of algorithm $\ref{alg:sample2}$ (conditional to the fact that the sample is not rejected).

\begin{algorithm}
\DontPrintSemicolon
 \caption{Get one sample of Counterfactual Disagreement \label{alg:sample2} }
 
     \SetAlgoLined
     Sample one banner from the dataset\;
     Reject it if there is not at least one clicked and one non clicked product\;
     Let $P_+$ the clicked product from this banner\;
     Reshuffle the banner by sampling an ordering from $\pi$.
     Define $P_-$ as the product at rank $r$ in the reshuffled banner.\;
     If $score(P_+) = score(P_-) $ , reject the sample\;
     Else return $\mathbf{1}_{score(P_+) < score(P_-)} $ \;
\end{algorithm}

This metric is obviously well defined only if $\pi$ is a non-deterministic policy (else $P_-$ would never be different from $P_+$, and all samples would get rejected). \\ 

It is straightforward to check that with this algorithm, $\mathbb{P}(P_+ = p_1, P_- = p_2, R = r ) > \mathbb{P}(P_+ = p_2, P_- = p_1 , R = r ) $  if and only if $\mathbb{P}( R=r |  P_r = p_1) > \mathbb{P}(R=r |  P_r = p_2) ) $,  where $P_r$ is the product placed at rank $r$. In other words we observe the pair $(P_+ = p_1, P_- = p_2 )$ at rank $r$ more often than the reversed pair if and only if $p_1$ has a higher click through rate than $p_2$ when placed at rank $r$. \\

Note that the expected value of this metric does still depend on the policy $\pi$ (because pairs of products are weighted by their propensity to appear at the same rank), but we argue that this dependency is much less a problem than the position bias of the pairwise disagreement.\\

Let's also note that in the special case when $\pi$ is a uniform distribution on orderings, ie when avoid the  position bias by uniformly shuffling the banner, our metric matches exactly the pairwise disagreement. Counterfactual disagreement can therefore be understood as a way to generalize pairwise disagreement to non uniformly shuffled banners.

\subsection{Interpretation: Recognizing which action lead to the reward}

Another way to define the metric is as follow: \\
Assume that when generating the data we collected two independent samples from $\pi$, and displayed only one of them randomly. If then we observe a click at rank $r$ on the displayed banner, can we retrieve which of those two banners was used with a model scoring the products ? \\
It should be more likely that the displayed sample was the one with the 'best' product at rank $r$. So a simple heuristic to recognize it is to pick the sample whose product at rank $r$ is scored highest by the scoring model. \\
Our metric is exactly the expected error rate of this 'banner recognition' scheme, conditioned on the fact that products at rank $r$ are different. \\
The comparison of the action which lead to the click with another sample from $\pi$ means that we cannot here only recognize the sample generated by $\pi$, but have to distinguish samples from $\pi$ and samples from the marginal distribution of samples of $\pi$ followed by a click.

\subsection{ Case of a Plackett-Luce distribution }

In algorithm~\ref{alg:sample2}, we need to be able to sample an ordering from distribution $\pi$.
In many practical cases however, the policy in production directly samples the set of recommended products and their ordering at the same time. 
The distribution  $\pi$ on orderings is then the conditional distribution knowing the set of products; and it may prove difficult to directly sample from this distribution.
\\
In particular, a commonly used distribution in this setting is the 'Plackett-Luce' distribution. It is defined as follow: the recommender system output some scores $x_1 , ... x_k$ for the k candidates products to display in the banner. Then products at position 1,2,3 ... are sampled successively, without replacement, by giving to each product a probability proportional to its score.
\\
We are not aware of any efficient way of sampling an ordering of the conditional distribution induced by Plackett-Luce when we know the set of items which was sampled. \\
But it is possible to compute, by dynamic programming, the conditional probability of putting a product $p$ at rank  $r$, knowing the set of displayed products. Appendix A details this method, with a complexity $O(n^2\times 2^n)$ with respect to the number $n$ of items in the banner. Noticing that algorithm~\ref{alg:sample2} actually only needs to get the product at rank $r$ in a sample from $\pi$, this method enables to compute our metric, at least for reasonably small values of $n$ (let's say less than 16, which is the case for most banners on Criteo data). But some more efficient methods to (approximately) sample from this distribution when $n$ grows larger would be of great interest here.

\subsection{Discussion on the variance of the metric}

If the policy $\pi$ is too close to a deterministic policy, most samples would get rejected in algorithm~\ref{alg:sample2}, leaving only few samples to estimate the expectation. This is an intrinsic limitation of the proposed metric.\\
We can also notice that if $\pi$ is almost deterministic on some context $x$ and more random on another context $x'$, then we would reject most samples on context $x$ while keeping many samples from context $x'$. Our metric thus downweights contexts where the policy is too deterministic. While not perfect (some context may be almost ignored in practice), we argue that this is a reasonable trade-off. It should be compared for example to the behavior of importance-weighting based metrics, which can typically get a huge variance from a few samples where the policy is too deterministic, and thus require in practice to use specific methods to control the variance.

\section{Experimental results on Criteo data }

At Criteo, we use a PLackett-Luce distribution to sample jointly the set and ordering in our banners. We also uniformly shuffle afterwards a small proportion of the banners. This enables us to compute ranking metrics, such as pairwise disagreement, without suffering from position bias of. However the small quantity of shuffled banners makes it more difficult to get significant results. Our metric can leverage all the non shuffled banners. \\

\begin{figure}
\includegraphics[width=\linewidth]{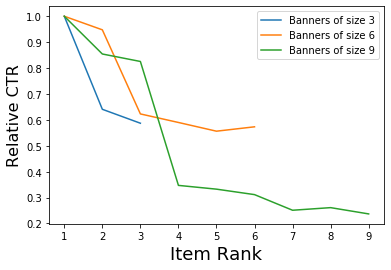}
\caption{\label{fig:bias_of_rank}  Position Bias on Criteo data }
\end{figure}

Figure~\ref{fig:bias_of_rank} shows the click through rate (CTR) as a function of the product rank on different subsets of banners from Criteo's data. We can observe that click through rate strongly depends on the rank, especially on large banners. We also noticed that this effect can be significantly modified by several parameters of the banner, such as its size, location on the page, ... 
This makes building a precise model of the effect of rank less trivial, and increases the usefulness of an offline metric not relying on such model.

\begin{figure}
  
\begin{subfigure}[h]{0.45\linewidth}
\includegraphics[width=\linewidth]{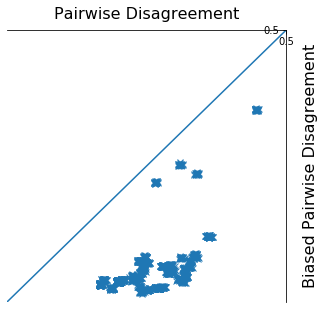}
\caption{\label{fig:correlation_1} With pairwise disagreement on non shuffled banners}
\end{subfigure}
\hfill
\begin{subfigure}[h]{0.45\linewidth}
\includegraphics[width=\linewidth]{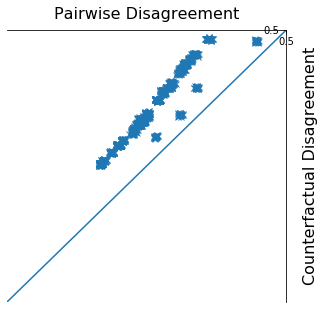}
\caption{\label{fig:correlation_2} With proposed metric}
\end{subfigure}%
\caption{\label{fig:correlation} Correlation of pairwise disagreement (shuffled banners)}
\end{figure}

We computed for about 40 models the pairwise disagreement and our proposed metric on non shuffled banners, and compared it to the pairwise disagreement on shuffled banners.\\
Figure \ref{fig:correlation} shows the collected data. We observe that pairwise disagreement is indeed severely affected by the position bias when using non shuffled banners, as can be seen by the weak correlation on figure~\ref{fig:correlation_1}. On the other hand our proposed metric \ref{fig:correlation_2} correlates reasonably well with pairwise disagreement of shuffled banners. It is also worth noting that the few outliers we can observe on the plot where found on models very far from our logging policy, which we were not going to test further anyway.\\
(As a side note, we also can also observe the value of both metrics seem to differ by an almost constant value. We think that this difference is caused mainly by some implementations details in our online randomization policy, which we can actually only approximate from offline data.)
\\
Because of the rejected samples, our metric has in practice more variance than the pairwise disagreement. On our data, it required between twice to thrice as many samples to get the same variance. But since we have only few shuffled banners, we could still get a large decrease of the variance compared to the pairwise disagreement on shuffled banners.\\
This seems to confirm that, at least in the Criteo use case, the metric we proposed in this article is a reasonable way to perform offline evaluation of our models, with a noise level significantly lower than what we have with the usual ranking metrics computed on shuffled banners.

\section{Variants and further work }
\subsection{ Resampling full banners instead of resampling only the ordering }

We defined here $\pi$ as the distribution on orderings. In practice however, the set of displayed products and their order are often sampled together, for example from a 'Plackett-Luce' distribution.\\ 
Instead, we could have defined $\pi$ to include the choice of the set of products and the ordering. (Thus replacing in algorithm \ref{alg:sample2}  "sample another ordering" by "sample another set of products and ordering".) \\
This choice would lead to another offline metric, but we do not know if it would be better correlated to online results.

\subsection{ Applying the same idea to other metrics than Pairwise Disagreement }

The proposed method seems quite straightforward to apply on the pairwise ranking loss, whose definition is very similar to pairwise disagreement, and could thus be used for learning.\\
The idea of recognising the displayed banner from a resample can also be extend it to other ranking metrics such as NDCG. It could also be used to define an offline metric for the case when we have a model scoring the full banner, instead of scoring the products separately, by comparing the score of the banner which lead to a click to the score of a resample from $\pi$.

\newpage

\appendix

\section{Implementation when products are sampled from a Plackett-Luce distribution }

\subsection{Notation}

Let $\mathcal{C}$ the set of candidate products we can display in some banner. \\
For a product $p \in \mathcal{C}$, let $score_p$ the score assigned to product $p$ by the model we used to build the banner.\\
For a subset $\mathcal{S} \in \mathcal{C}$, we note $scores_\mathcal{S} := \sum\limits_{ p \in \mathcal{S} } score_p $ the sum of scores of products of $\mathcal{S}$. \\

Finally, we define the following random variables:\\
$P_k$ the product displayed at rank $k$ in the banner \\
$D_k := \{ P_1, ... P_k \}$ the set of the first $k$ products.\\
$D := D_n$ the set of displayed products.\\

\subsection{the Plackett-Luce distribution }

In practice at Criteo, we sample jointly the set $D$ of displayed products and their ordering, from a 'Plackett-Luce' distribution. \\
This sampling is implemented as follow: we draw the products one by one without replacement from the set $\mathcal{C}$ of candidates, assigning to each product a weight proportional to its score.
In other words, the probability of choosing product $p$ at rank $i$ after selecting items $p_1,... p_{i-1}$ is defined as:
\begin{equation} \label{eq:proba_conditional_at_rank_i}
 \mathbf{P}( P_i = p | P_1 = p_1, ... P_{i-1} = p_{i-1}  ) = \frac{ score_p \times \mathbf{1}_{ p \not\in \{p_1 ... p_{i-1} \}  }  }{ scores_\mathcal{C} - scores_{\{p_1 ... p_{i-1} \}}  }  
 \end{equation}

The probability of displaying banner $(p_1, ... p_n)$ is then:

\begin{equation} \label{eq:proba_banner}
\mathbf{P}( P_1=p_1, ... P_n=p_n ) = \frac{ \prod\limits_{ i=1 }^n score_{p_i}  }{ \prod\limits_{ i=1 }^n ( \sum\limits_{p \in \mathcal{C}} score_p - \sum\limits_{k=1}^i score_{ p_k }  )      } 
\end{equation}

\subsection{Induced distribution on orderings}

Let $\mathcal{D} := \{ p_1 , ... p_n \} $ the set of products in one sampled banner. ($p_1 , ... p_n$ are thus realizations of the random variable $P_1,... P_n$ ) \\

The policy $\pi$ used in \ref{alg:sample2} is the distribution on orderings of the products $p_1, ... p_n$, conditioned by the fact that $D$ is the set $\{p_1,... p_n\}$.
The set of ordering of those products can be identified with the set $\mathlarger{\mathlarger{\mathlarger{\sigma_n}}}$ of permutations of $(p_1, p_n)$.\\
For $\sigma \in \mathlarger{\mathlarger{\mathlarger{\sigma_n}}}$, $\pi(\sigma)$ is thus defined as:
\begin{align}
 \label{eq:def_pi}
\pi(\sigma) &:= \mathbb{P}( P_1 = p_{sigma(1)},... P_n = p_{sigma(n)} |  D = \mathcal{D} ) \\
            &=  \frac{ \mathbb{P}( sigma ) }{ \sum\limits_{ \omega \in \mathlarger{\mathlarger{\mathlarger{\sigma_n}}} }   \mathbb{P}( \omega )  } \\
\end{align}
where we noted $\mathbb{P}( \sigma ) :=  \mathbb{P}( P_1 = p_{\sigma(1)},... P_n = p_{\sigma(n)}) $.\\

\subsection{Sampling from the induced distribution}

To implement \ref{alg:sample2}, we need to sample from $\pi$ and find the product at rank $r$ in this sample.\\
One naive method would be to explicitly use equation \ref{eq:def_pi} to compute the probability of each of the orderings. Obvious limitations is that there are $n!$ such orderings, making it prohibitively costly even for small values of $n$. (In most of Criteo data, $n$ varies between $2$ and $16$)\\
Another naive implementation would be sampling from the full set of candidates following equation \ref{eq:proba_banner}, and reject each banner whose set of products does not match the displayed set. But the probability of not rejecting the sample is usually too low to make this approach practical. \\
Actually, we are not aware of any efficient way to sample $\sigma$ from the induced distribution. But let us notice that \ref{alg:sample2} only require the product at rank $r$ in samples from $\pi$. 

\subsection{Probability of getting product p at rank $r$ in a sample of $\pi$}

We show here how to compute, for each of the $n$ products $p \in \mathcal{D}$, the probability $\mathbb{P}(P_r = p | D=\mathcal{D})$ that this product placed at rank $r$ in a sample of $\pi$, with a complexity only $O(n^2 \times 2^n)$.
While still unpractical for large banners, it is reasonable to use for the typical values of $n$ in our banners, and enables to implement \ref{alg:sample2} by sampling directly the product at rank $r$. \\

Lemma 1:\\
For any non empty set $\mathcal{S} \subset \mathcal{D}$ of size $k$:
$$\mathbb{P}(D_k = \mathcal{S})  = \sum\limits_{ p \in s} \mathbb{P}( D_{k-1} = \mathcal{S}_{\setminus\{p\}}  ) \cdot \frac{ score_p }{ scores_{\mathcal{C}} - scores_{\mathcal{S}_{\setminus\{p\}} } } $$
Indeed, if $k$ is the size of $S$, the event 'the first $k$ sampled items are the element of $S$' is the disjoint union of the events 'the first $k-1$ sampled items are the element of $S\setminus p$, and the next item is $p$', and $\frac{ score_p }{ scores_{\mathcal{C}} - s_{S\setminus_p } }$ is exactly probability of sampling $p$ as the next item when we just sampled the other elements of $S$.\\

With the convention $\mathbb{P}( D_0 = \varnothing ) = 1 $, we can use lemma 1 to compute $\mathbb{P}(D_{ size_k } = s)$ for each $S \subset $. There are $O(2^n)$ such sets, and by using a cache of the results, each of them requires to iterate on at most $n$ products. The complexity here is thus $O(n \times 2^n)$. \\

Lemma 2:\\
Let $\mathcal{S} \subset \mathcal{D}$ of size at least $k$ with  $k \geqslant r$ , and $p_0 \in S$.\\
If $ k = r $, then:
$$\mathbb{P}(D_k=\mathcal{S},P_r = p_0) = \mathbb{P}(D_k=\mathcal{S}_{\setminus\{p_0\}} ) \cdot \frac{ score_{p_0} }{ scores_{\mathcal{C}} - scores_{\mathcal{S}_{\setminus\{p_0\}} } } $$
Else:
\begin{multline*}
\mathbb{P}(D_k=\mathcal{S},P_r = p_0) =\\
    \sum\limits_{ p \in \mathcal{S}, p \neq p_0 } \mathbb{P}( D_{k-1} = \mathcal{S} \setminus_p ,P_r = p_0 ) \cdot \frac{ score_p }{ scores_{\mathcal{C}} - scores_{\mathcal{S}_{\setminus\{p\}} } } 
\end{multline*}

The first case comes from the fact that for a set of size $r$, the event 'first sampled items are elements of S, and $p_0$ is at position $r$' is equivalent to 'first sampled items are elements of $\mathcal{S}$ except $p_0$, and the next item is $p_0$.\\
The second case follows in a similar way by partitioning the event 'first sampled items are elements of S, and $p_0$ is at position $r$' as the disjoint union with respect to the last sampled product $p$.\\

Using the precomputed results from lemma 1, We can now use lemma 2 to compute $\mathbb{P}(D_{size(k)}=\mathcal{S},P_r = p_0) $ for each subset $\mathcal{S}$ of size at least $r$, and each element $p0$. Complexity is here now $O(n^2 \times 2^n)$, because we should also iterate on element $p_0$. \\

In particular, this enables us to get the value of $\mathbb{P}(D=\mathcal{D},P_r = p)$ and $\mathbb{P}(D=\mathcal{D})$, which finally enables us to compute $\mathbb{P}(P_r = p | D=\mathcal{D})$ with the Bayes rule.